\documentclass[preprint,12pt]{elsarticle}

\usepackage{lineno,hyperref}
\modulolinenumbers[5]

\usepackage{algorithm,algorithmic}
\usepackage{caption,subfig}
\usepackage{amssymb,amsmath}
\usepackage{pifont}
\usepackage{todonotes}
\usepackage{color}
\usepackage{xspace,array}
\usepackage{setspace}
\usepackage{listings}

\journal{Engineering Applications of Artificial Intelligence}
\date{February 13, 2019}

\usepackage{fancyhdr}
\biboptions{authoryear}

\begin{document}

\begin{frontmatter}

\title{JCLEC-MO: a Java suite for solving many-objective optimization engineering problems}

\author{Aurora Ram\'irez}
\ead{aramirez@uco.es}
\author{Jos\'e Ra\'ul Romero\corref{corauthor}}
\ead{jrromero@uco.es.}
\author{Carlos Garc\'ia-Mart\'inez\corref{x}}
\ead{cgarcia@uco.es}
\author{Sebasti\'an Ventura\corref{x}}
\ead{sventura@uco.es}
\address{Department of Computer Science and Numerical Analysis, University of C\'ordoba, 14071 C\'ordoba Spain}
\cortext[corauthor]{Corresponding author. Tel.: +34 957 21 26 60}

\begin{abstract}
Although metaheuristics have been widely recognized as efficient techniques to solve real-world optimization problems, implementing them from scratch remains difficult for domain-specific experts without programming skills. In this scenario, metaheuristic optimization frameworks are a practical alternative as they provide a variety of algorithms composed of customized elements, as well as experimental support. Recently, many engineering problems require to optimize multiple or even many objectives, increasing the interest in appropriate metaheuristic algorithms and frameworks that might integrate new specific requirements while maintaining the generality and reusability principles they were conceived for. Based on this idea, this paper introduces JCLEC-MO, a Java framework for both multi- and many-objective optimization that enables engineers to apply, or adapt, a great number of multi-objective algorithms with little coding effort. A case study is developed and explained to show how JCLEC-MO can be used to address many-objective engineering problems, often requiring the inclusion of domain-specific elements, and to analyze experimental outcomes by means of conveniently connected R utilities.

\end{abstract}


\end{frontmatter}


\section{Introduction}\label{sec:introduction}

Optimization problems frequently appear in the engineering field, but their characteristics make the application of mathematical methods not always feasible~\citep{Singh16}. Hence, the use of efficient search methods has experienced a significant growth in the last years, specially for those engineering problems where there are multiple objectives that require to be simultaneously optimized~\citep{Marler04}. A recurrent situation in engineering is the need of jointly optimizing energy consumption, cost or time, among others. All these factors constitute a paramount concern to the expert, and represent conflicting objectives, each one having a deep impact on the final solution~\citep{Marler04}. Initially applied to single-objective problems, metaheuristics like evolutionary algorithms (EAs) have been successfully applied to the resolution of multi-objective problems (MOPs) in engineering, such as the design of efficient transport systems~\citep{Dominguez14} or safe civil structures~\citep{Zavala13}.

The presence of a large number of objectives has been recently pointed out as an intrinsic characteristic of engineering problems~\citep{Singh16}, for which the currently applied techniques might not be efficient enough. It is noteworthy that other communities are also demanding novel techniques to face increasingly complex problems, what has led to the appearance of the many-objective optimization approach~\citep{Lucken14,Li15}. This variant of the more general multi-objective optimization (MOO) is specifically devoted to overcome the limits of existing algorithms when problems having 4 or more objectives, known as many-objective problems (MaOPs), have to be faced. Even though each metaheuristic follows different principles to conduct the search, their adaptation to deal with either MOPs or MaOPs share some similarities, such as the presence of new diversity preservation mechanisms or the use of indicators~\citep{Li15,Mishra15}. The resulting many-objective algorithms have proven successful in the engineering field too~\citep{Li14,LopezJaimes14,Cheng17}, where specialized software tools have begun to appear~\citep{Hadka15}.

In fact, the availability of software suites is one of the factors that most influences engineers when selecting a solution or algorithm~\citep{Marler04}, as they can greatly reduce coding efforts and even provide some guidance to engineers. In this context, metaheuristic optimization frameworks (MOFs)~\citep{Parejo12} seem to go one step further, as they may integrate environments not only providing a collection of algorithms or code templates, but also other general utilities to properly configure them and analyze outputs. MOFs are modular and adaptable in different ways, and should enable the introduction of specific domain knowledge and constraints in terms of the representation and evaluation of solutions~\citep{LopezJaimes14,Singh16}.

Focusing on the resolution of MOPs, these suites are expected to keep the principles of multi-objective optimization by making the appropriate adaptations for their components to deal with multiple objectives. At the same time, MOFs still need to consider aspects like efficiency, utility and integrability if a broad industrial adoption is sought. Among the currently available alternatives, there are some specialized frameworks like jMetal~\citep{Durillo11} and MOEA Framework~\citep{Hadka17}, whose main strength lies on a more extensive catalog of recent algorithms. Besides, other general-purpose MOFs like ECJ~\citep{White12}, HeuristicLab~\citep{Elyasaf14} or JCLEC~\citep{Ventura08} benefit other aspects like their ease of use and greater availability of components to represent and modify the solutions are their key advantages.

A mix of both alternatives would enable to take advantage of reusability, maturity and the reduction of the learning curve promoted by general-purpose components, whereas specialization might bring the suite closer to comply with current requirements of industry. At this point, JCLEC has been reported as a competitive tool due to its large number of customizable components, which can be combined to solve user-defined optimization problems~\citep{Parejo12}. In addition, JCLEC can be easily integrated with other systems because of its regular use of standards like XML. Its core elements are defined at a high level of abstraction, providing the required flexibility to build new functionalities on top of a stable platform. Therefore, JCLEC has become an interesting baseline MOF to be extended to adopt different metaheuristics for the resolution of both MOPs and MaOPs within an industrial environment.

To this end, this paper presents JCLEC-MO, an extensible framework providing suitable search elements and techniques for multi- and many-objective optimization. The preliminary architecture~\citep{Ramirez15}, only focused on multi-objective evolutionary algorithms (MOEAs), has been refined and significantly evolved to include new types of algorithms and support for other metaheuristics. As a result, JCLEC-MO provides generic metaheuristic models that have been conveniently adapted to the precepts of MOO, and still preserves the valuable characteristics of a general-purpose solution. The conceptual algorithmic model proposed to achieve independence and a significant scalability is a distinctive characteristic of JCLEC-MO. It is also competitive in terms of the available catalog of algorithms, mechanisms to assess their performance and reporting capabilities. A case study shows how this suite enables the resolution of a many-objective engineering problem, thus serving to illustrate how user-defined components should be conceived and how the returned solutions could be analyzed, e.g. by using R functionalities.

The rest of the paper is organized as follows. Section~\ref{sec:background} provides an essential background on metaheuristics for multi- and many-objective optimization and MOFs. Existing frameworks for solving multi-objective problems are analyzed in Section~\ref{sec:related}. Section~\ref{sec:design} presents the design criteria and architecture of JCLEC-MO. A more detailed description of the software functionalities and its modular organization is provided in Section~\ref{sec:modules}. Then, Section~\ref{sec:example} develops an illustrative case study to show the applicability and use of JCLEC-MO as a supportive tool for engineers. A discussion of the benefits of JCLEC-MO compared to other available MOFs is presented in Section~\ref{sec:comparison} and, finally, conclusions are outlined in Section~\ref{sec:conclusions}.

\section{Background}\label{sec:background}

Metaheuristics, just like evolutionary algorithms~\citep{Eiben15} and particle swarm optimization (PSO)~\citep{Poli07}, are well-known techniques to address optimization problems due to their efficiency and independence of the problem formulation. Based on the principles of natural evolution, EAs manage a set of candidate solutions (population of individuals) that are iteratively selected, recombined and mutated to gradually produce improved solutions. In PSO, each particle represents a potential solution that changes its position and velocity influenced by the rest of particles. Other bio-inspired metaheuristics imitate the behavior of other forms of living beings, such as ants or bees, when looking for resources like food sources~\citep{Boussaid13}.

These paradigms were promptly adapted to deal with problems having more than one objective. Solving a MOP involves finding the values of a group of decision variables that jointly optimize a set of objective functions, while satisfying other possible constraints~\citep{Coello07}. In this scenario, optimal solutions, a.k.a. non-dominated solutions, are those for which there is no other feasible solution with better outcomes for all the objectives, so they reach the best trade-off among them. Therefore, multi-objective algorithms aim at obtaining a good approximation to the Pareto front (PF), namely the set of non-dominated solutions within the objective space. Finally, the expert will make the choice.

It is worth remarking that, due to the specific characteristics of a MOP, any multi-objective metaheuristic needs to reconsider three main search concepts: fitness assignment, diversity preservation and elitism~\citep{Talbi09}. Therefore, dominance rankings, the definition of diversity measures or the creation of an external archive to promote elitism are examples of mechanisms that frequently appear in MOEAs~\citep{Konak06}. Algorithm~\ref{alg:moeacode} shows how these elements are integrated in the general structure of a MOEA. Furthermore, given that they worked well in MOEAs and were not dependent on how the algorithm creates or modifies solutions, they were subsequently adopted by other metaheuristics~\citep{Coello07}. For instance, multi-objective PSO includes a sort of mutation, named turbulence, to promote diversity, and the set of leaders, i.e. the best particles, are selected according to dominance principles and kept within an external archive~\citep{Reyes06}.

\begin{algorithm}[!t]
\caption{Pseudocode of a MOEA (adapted from~\citep{Coello07})}
\label{alg:moeacode}
\begin{algorithmic}[1]
	\STATE {Initialize population $P$ and archive $P^*$}
	\STATE {Evaluate objective functions over $P$}
	\STATE {Assign fitness to $P$ based on dominance and diversity}
	\WHILE {not stopping condition (e.g. number of generations)}
		\STATE {Selection of parents: $P^i \leftarrow select(P \cup P^*)$ }
		\STATE {Recombination and mutation of individuals: $P^{ii} \leftarrow genOps(P^i)$}
		\STATE {Evaluate objective functions over $P^{ii}$}
		\STATE {Assign fitness to ($P\cup P^{ii}$) based on dominance and diversity}
        \STATE {Replace $P$ choosing from ($P\cup P^{ii}$)}
		\STATE {Update archive: $P^* \leftarrow update(P,P^{ii},P^*)$}
	\ENDWHILE
\end{algorithmic}
\end{algorithm}

In the last years, researchers have stressed the need of applying metaheuristics to solve many-objective problems, a term commonly accepted in the literature for those having 4 or more objectives~\citep{Zhou11,Lucken14}. First attempts to address MaOPs were focused on adapting already existing evolutionary algorithms for MOPs~\citep{Adra11}. However, other specific mechanisms have appeared more recently, such as the use of indicators or reference points to guide the search~\citep{Li15}. At present, many-objective approaches---originally integrated into EAs---can be found in conjunction with other metaheuristics like ant colony optimization~\citep{Falcon17}, bee colony optimization~\citep{Luo17} or PSO~\citep{Figueiredo16}.

The extensive application of metaheuristics to real-world complex problems is also reflected in the area of engineering. The intrinsic characteristics of these problems, which are affected by multiple decision factors and constraints and may require time-consuming simulations to evaluate solutions, make MOEAs specially appealing~\citep{Zhou11}. Other paradigms are beginning to draw more attention in the last years. For instance, Zavala \textit{et al.}~\citep{Zavala16} conducted a comparative study of several multi-objective metaheuristics to improve the design of cable-stayed bridges. Similarly, examples of the use of many-objective metaheuristics for aiding engineers in a variety of areas can be found in the literature, such as vehicle control systems~\citep{Cheng17} (7 objectives) or the design of airfoils~\citep{LopezJaimes14} (6 objectives). A PSO algorithm was also proposed for this latter domain with 5 objectives~\citep{Wickramasinghe10}, as well as for the balance of risk and performance objectives in wind-sensitive structures~\citep{Li14}.

Despite their wide popularity in academic environments, engineers from an industrial context could find difficult to work with metaheuristics without any kind of tool support. To mitigate the skill gap, MOFs can act as a bridge between the research in the field of optimization and its adaptation to the needs of the engineering industry. Notice that metaheuristic optimization frameworks do not only provide most of the components taking part in the search algorithm, i.e. solution encodings, operators, selection mechanisms and iterative processes, but they also provide the required support to create experiments, monitor their execution and report outcomes~\citep{Parejo12}. MOFs are specially well-suited for non-expert users, thus facilitating the selection and customization of components, mainly with the challenges to come with the increasing complexity and number of objectives being considered. 

MOFs are also valuable suites for developing and verifying new proposals. Their modularity make code more reusable and, consequently, the required development and testing efforts can be reduced. However, MOFs need to maintain a modular design, provide clear guidelines and promote extensibility. These aspects are significantly more relevant for multi- and many-objective metaheuristics, since each paradigm may propose its own procedure to conduct search, while they could still use the same mechanisms to deal with MOPs and MaOPs.

\section{Related work}\label{sec:related}

In recent years, the number of available MOFs has grown, possibly motivated by the existence of different audiences with particular purposes. Parejo \textit{et. al}~\citep{Parejo12} compared a selected group of frameworks and evaluated their characteristics, specially those referred to the diversity of techniques, their level of customization and the quality of the documentation. However, the potential to address MOPs was barely considered, as it was mostly focused on the availability of certain algorithms proposed in early 2000s. In this section, we firstly present an updated list of general-purpose frameworks including some kind of support for MOO. Next, a more detailed analysis of MOFs specially for MOO is discussed.

\subsection{General-purpose MOFs}\label{subsec:generalMOFs}

Table~\ref{tab:mofs1} categorizes the best-known general-purpose frameworks with respect to the following aspects: the list of supported metaheuristic paradigms, either for single- or multi-objective optimization (or both); types of encoding available, what also serves to a certain extent to demonstrate the variety of problems that they could address; the way in which optimization problems are defined (to be minimized or maximized, with or without constrains); the specific multi-objective algorithms currently supported; the set of multi-objective benchmarks and the collection of assessment metrics, a.k.a. quality indicators. 

\begin{table}[!t]
\begin{center}
\caption{Summary of the characteristics of general-purpose MOFs}
\scalebox{0.75}{
\begin{tabular}{|>{\centering\arraybackslash}m{3cm}|>{\centering\arraybackslash}m{4cm}|>{\centering\arraybackslash}m{4cm}|>{\centering\arraybackslash}m{4cm}|} \hline
\textbf{Characteristic}		& \textbf{ECJ v25 (2017)} & \textbf{HeuristicLab v3.3.15 (2018)} & \textbf{EvA v2.2 (2015)} \\\hline
\hline
Metaheuristics				& DE, EDA, ES, GA, GE, GP, PSO & ES, GA, GE, GP, LS, PSO, SS, TS, SA, VNS & DE, EP, ES, GA, GP, HC, PSO, SS, SA \\\hline

Encodings					& binary, integer, real, tree & binary, integer, real, tree & binary, integer, real, tree \\\hline

Optimization problems		& c/u, min/max & u, min/max & c/u, min \\\hline

MOO algorithms				& NSGA-II, SPEA2 & MO-CMAES, NSGA-II & MO-CMAES, MOGA, NSGA, NSGA-II, PESA, PESA-II, Random Weight GA, SPEA, SPEA2, VEGA \\\hline

MOO benchmarks				& Fons.\&Flem., Kursawe, Poloni, Quagli.\&Vicini, Schaffer, Sphere, ZDT & Fonseca, Kursawe, Schaffer, DTLZ, ZDT & TF \\\hline

Quality indicators			&  & GD, HV, Spacing & ER, GD, HV, Max. PF error, ONVG\\\hline
\hline
\textbf{Characteristic}		& \textbf{Opt4J v3.1.4 (2015)} & \textbf{PaGMO v2.6 (2017)} & \textbf{JCLEC v4 (2014)}\\\hline
\hline
Metaheuristics				& DE, GA, PSO, SA & ABC, DE, ES, GA, PSO, SA & GA, GP\\\hline

Encodings					& binary, integer, real & integer, real, mixed & binary, integer, real, tree\\\hline

Optimization problems		& c/u, min/max &  c/u, min & min/max\\\hline

MOO algorithms				& NSGA-II, SPEA2, SMS-EMOA, OMOPSO & MOEA/D, NSGA-II & NSGA-II, SPEA2\\\hline

MOO benchmarks				& DTLZ, Knapsack, LOTZ, Queens, WFG, ZDT & DTLZ, ZDT & \\\hline

Quality indicators			& HV & HV &\\\hline

\multicolumn{4}{c}{\footnotesize{ABC: artificial bee colony, DE: differential evolution, EDA: estimation of distribution algorithms}}\\
\multicolumn{4}{c}{\footnotesize{EP: evolutionary programming, ES: evolution strategy, GA: genetic algorithm}}\\
\multicolumn{4}{c}{\footnotesize{GE: grammatical evolution, GP: genetic programming, HC: hill climbing, LS: local search}}\\
\multicolumn{4}{c}{\footnotesize{SS: scatter search, SA: simulated annealing, TS: tabu search, VNS: variable neighborhood search}}\\
\multicolumn{4}{c}{\footnotesize{c: constrained, u: unconstrained, min: minimization, max: maximization}}
\end{tabular}
}
\label{tab:mofs1}
\end{center}
\end{table}

Firstly, ECJ is a well-known Java-based research system for evolutionary computation~\citep{White12,Luke17}, for which the different steps of MOEAs, i.e. selection, evaluation and replacement, are implemented separately from the rest of the search process established in the breeding pipeline. This highly modular design was followed to deploy the two MOEAs provided, SPEA2 and NSGA-II~\citep{Coello07}, whose performance can be assessed against a variety of test functions. On the other hand, developed in C\# for Microsoft .NET, HeuristicLab~\citep{Elyasaf14,Wagner14} provides a fully functional environment with a user graphical interface to run diverse optimization algorithms, enabling the representation and evaluation of MOPs. Additionally, some benchmarks have been also included. Analogously to ECJ, only two MOEAs are available, which have the same structure than any single-objective algorithm.

Other general-purpose Java libraries have also gain attention for dealing with MOO requirements, being able to facilitate a modular design of these algorithms. For instance, MOEAs in EvA~\citep{Kronfeld10} are declared from a generic class, named \emph{MultiObjectiveEA}, which has to be configured jointly with an optimization strategy, an archiver and an information retrieval strategy. The last two elements are defined by the specific MOO approach, providing up to 10 different MOEAs (see Table~\ref{tab:mofs1}). This suite has also the most complete catalog of quality indicators, but just one benchmark for tests. On the other hand, Opt4J~\citep{Lukasiewycz11} includes multi-objective implementations for EAs and PSO. These approaches are developed on the basis that the selection and replacement steps are both defined by the interface \emph{Selector}. In Opt4J, selectors are viewed as configurable elements of an optimizer that separately control the rest of the search. This suite also includes implementations of popular benchmarks, for both continuous and combinatorial tests.

Likewise, PaGMO/PyGMO~\citep{Biscani10,Izzo12} is a C++/ Python platform that provides the necessary support to build parallel global optimization algorithms. Two MOEAs are currently available, though they apparently suffer from a lack of customization capacity since the definition of genetic operators is embedded within the algorithm itself. Finally, JCLEC is a highly modular framework for evolutionary computation written in Java. Like most of these frameworks, the current version of JCLEC is mostly conceived to address single-objective optimization problems, even when it implements the two most usual MOEAs. Nevertheless, one key factor of JCLEC is its extensibility, as its core elements are independent of each other, defined at a high level of abstraction, and the interface and object specification is clear and well structured. Hence, it does not only provide an appropriate stable platform to build new MOO functionalities, but also offer the flexibility required to integrate new metaheuristics.

\subsection{MOO-specific MOFs}\label{subsec:specificMOFs}

Table~\ref{tab:mofs2} shows the most popular frameworks specifically oriented to the resolution of MOPs. It is worth remarking that, in this case, algorithms, benchmarks and quality indicators acquire more importance.

\begin{table}[!t]
\begin{center}
\caption{Summary of MOO-specific MOFs and their characteristics}
\scalebox{0.8}{
\begin{tabular}{|>{\centering\arraybackslash}m{3cm}|>{\centering\arraybackslash}m{6cm}|>{\centering\arraybackslash}m{6cm}|} \hline
\textbf{Characteristic}	& \textbf{PISA (2009)} & \textbf{ParadisEO-MOEO v2.0.1 (2012)}  \\\hline
\hline
Metaheuristics					& ES, GA & HC, ILS, GA, PSO, SA, TS, VNS \\\hline

Encodings						& binary, real & binary, integer, real \\\hline

Optimization problems			& u, min & u, min/max \\\hline

MOO algorithms					& $\epsilon$-MOEA, FEMO, HypE, IBEA, MSOPS, NSGA-II, SEMO2, SHV, SPAM, SPEA2 & DMLS, IBEA, IBMOLS, MOGA, NSGA, NSGA-II, PLS, SEEA, SPEA2 \\\hline

MOO benchmarks					& BBV, DTLZ, Knapsack, MLOTZ, WFG & Schaffer and external contributions (DTLZ, Flowshop, WFG, ZDT) \\\hline

Quality indicators				& $I_{\epsilon}$, $I_{\epsilon+}$, HV, R2, R3 & $I_{\epsilon+}$, contribution, HV \\\hline
\hline
\textbf{Characteristic}	& \textbf{jMetal v5.4 (2017)} & \textbf{MOEA Framework v2.12 (2017)} \\\hline
\hline
Metaheuristics					& CE, CRO, DE, ES, GA, LS, PSO, SS & DE, ES, GA, GP, GE, PSO \\\hline

Encodings						& binary, integer, real, mixed & binary, integer, real, tree \\\hline

Optimization problems			& c/u, min & c/u, min\\\hline

MOO algorithms					& AbYSS, CellDE, dMOPSO, GDE3, GWASF-GA, IBEA, MOCell, MOEA/D, MOEADD, MOCHC, MOMBI, MOMBI-II, NSGA-II, NSGA-III, OMOPSO, PAES, PESA2, R-NSGA2, SMPSO, SMS-EMOA, SPEA2, WASF-GA & DBEA, $\epsilon$-MOEA, GDE3, IBEA, MO-CMAES, MOEA/D, MSOPS, NSGA-II, NSGA-III, OMOPSO, PAES, PESA2, RVEA, SMS-EMOA, SMPSO, SPEA2, VEGA\\\hline

MOO benchmarks					& CDTLZ, CEC'09, DTLZ, Fonseca\&Fleming, GLT, Kursawe, LGZ, LZ, Schaffer, TSP, WFG, ZDT, among others & CDTLZ, CEC'09, DTLZ, Fonseca\&Fleming, GLT, Knapsack, Kursawe, LOTZ, LZ, Poloni, Schaffer, WFG, ZDT, among others\\\hline

Quality indicators				& $I_{\epsilon}$, ER, $\Delta S$, GD, HV, IGD, IGD+, R2, spread, two set coverage & $I_{\epsilon+}$, contribution, GD, HV, IGD, max. PF error, R1, R2, R3, spacing, two set coverage\\\hline
\multicolumn{3}{c}{\small{CE: cellular algorithms, CRO: coral reefs optimization, ILS: iterated local search}}\\
\end{tabular}
}
\label{tab:mofs2}
\end{center}
\end{table}

PISA and ParadisEO-MOEO could be considered the two first attempts to provide specific suites for MOO. Even though both projects were updated for the last time several years ago, they laid some foundations for future proposals. On the one hand, PISA~\citep{Bleuler03} was conceived as a language-independent platform based on file interchangeability, where a monitor is in charge of connecting the algorithm, named selector, to a variator used to define all the domain-specific elements. On the other hand, ParadisEO is a multi-purpose MOF, developed in C++, with a specific module for MOO~\citep{Liefooghe11}. This suite proposed an interesting conceptual model for an algorithm, in which the fitness assignment, the selection method and the replacement strategy are independent to promote modularity. This model has been implemented by popular evolutionary algorithms like NSGA-II and SPEA2, as well as some other in-house developments of local search methods (DMLS, IBMOLS and PLS).

jMetal~\citep{Durillo11,Nebro15} is the first Java framework specifically conceived for MOO, and still one of the most active projects nowadays. This suite consists of a library that provides a large number of multi-objective algorithms based on different metaheuristic techniques, including different types of evolutionary algorithms, such as DE, SS and GA, and PSO. Internally, two abstract classes, namely \emph{AbstractEvolutionaryAlgorithm} and \emph{AbstractParticleSwarmOptimisation}, decompose the corresponding iterative process into different steps. It is worth noting that jMetal does not explicitly separate the metaheuristic model from its specific adaptations for MOO. Next to this, MOEA Framework~\citep{Hadka17} is another Java-based suite that combines highly-modular native implementations (see Table~\ref{tab:mofs2}) with the possibility of running other non-native algorithms from PISA and a former version of jMetal. Consequently, the resulting catalog of algorithms is most extensive among the MOO-based alternatives. Nevertheless, it should be considered that each algorithm may present a different structure and level of customization, what also occurs with native implementations of EAs and PSO algorithms. The collection of benchmarks and quality indicators provided by both frameworks is particularly extensive, as shown in Table~\ref{tab:mofs2}.


\section{Architecture of JCLEC-MO}\label{sec:design}

On the basis of the lessons learned from existing MOFs, this section introduces the identified design principles and construction guidelines for JCLEC-MO. More specifically, Table~\ref{tab:design-principles} collects the principles to be applied, with a clear focus on the resolution of user-defined problems, a correctly structured modularization and the independence between components in order to promote extensibility and usability. Other aspects like integration facilities and external access to analytical functionalities also require precise specifications.

\begin{table}[!t]
\begin{center}
\caption{Design principles of JCLEC-MO}
\scalebox{0.92}{
\setstretch{1.0}
\small
\begin{tabular}{|p{4cm}|p{10cm}|} \hline
\textbf{Design principle}	&\textbf{Description}\\\hline
Generality preservation		&
Generality principles for this kind of software~\citep{Gagne06,Ventura08} should be preserved. Any design decision has been made according to the precepts of best practices of software engineering, e.g. use of design patterns, as a way to guarantee modularity and extensibility. Extension mechanisms have also been clearly specified.
\\\hline

Design extensibility		&
With the aim of promoting adaptability, scalability and reusability among different metaheuristics, there should be a clear distinction between the core elements of the metaheuristic models and those parts specifically adapted to the resolution of MOPs and MaOPs. 
\\\hline

Updated availability		&
A competitive collection of well-known multi- and many-objective algorithms based on EAs and PSO is desirable. With adequate specifications and mechanisms for extensibility and modularity, end-users should be able to easily enlarge the collection by themselves.
\\\hline

Independence of problem definition &
Algorithms should be modular enough to work with the minimal set of restrictions concerning the number and mathematical formulation of objectives, the solution encoding and the presence of constraints. 
\\\hline

Domain adaptability			&
The configuration of domain-specific elements referred to the optimization problem should be assumable, either for those selected from the catalog of elements provided by the framework or for those provided by third-parties as external contributions.
\\\hline

Batch processing and parallel evaluation	& 
Experiments should be defined as a sequence of processes, automatically executed, where solutions could be evaluated in parallel.
\\\hline

Experimental support		&
Flexible and understandable mechanisms for configuring experiments, reporting capabilities and outputs should be available, as well as the collection of usual benchmarks and quality indicators.
\\\hline

Interoperability and standardization	&
Design decisions should ensure the interoperability between procedures and tools provided by third-parties, such as algorithms or analytical tools, as well as the compliance with standards related to data formats and communication interfaces.
\\\hline
\end{tabular}
}
\label{tab:design-principles}
\end{center}
\end{table}

\begin{figure*}[!t]
	\centering
		\includegraphics[width=\textwidth]{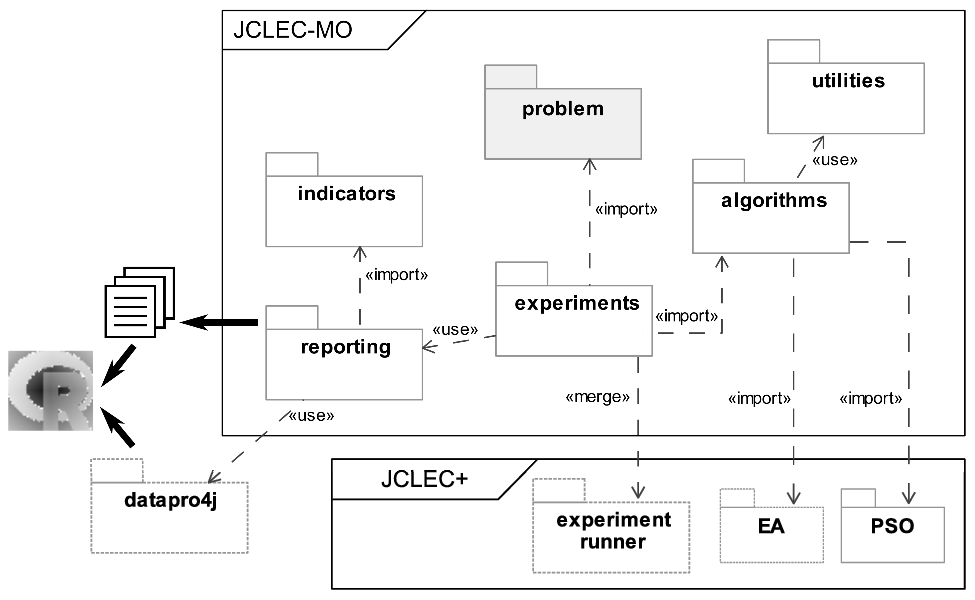}
	\caption{Overall module organization of JCLEC-MO}
	\label{fig:blocks}
\end{figure*}

As a suite, JCLEC-MO is built on top of the framework JCLEC, as a way to integrate and reuse ground elements. Nevertheless, JCLEC-MO still keeps its own identity by proposing a new architecture to comply with the expected MOO-specific requirements. Following the precepts of software design, Figure~\ref{fig:blocks} displays the overall organization of JCLEC-MO, in terms of its core components and their mutual interactions. Dashed lines state for external libraries, and the shaded box contains those modules demanding some user-provided code. 

As can be seen, the framework is made up of three main elements. Firstly, JCLEC-MO is the largest module, in charge of implementing MOO-specific functionalities, including the algorithms for MOO. Internally, the block \textit{experiments} is responsible for connecting the rest of components in this layer, as well as controlling the execution of \textit{algorithms} by operating on the configuration and running capabilities of JCLEC. The domain-specific elements conceived to solve an optimization \textit{problem}, i.e. objective functions or new operators, need to be externally defined by the end-user. To this end, a clear specification of interfaces and abstract classes is provided as a reference. Furthermore, other general \textit{utilities} enable a more rapid development and testing of new algorithms by encapsulating basic operations in the multi-objective space and making benchmarks available.

Secondly, JCLEC+ is an extension of JCLEC, specially conceived and developed to serve as a bridge between JCLEC and JCLEC-MO. This module provides basic mechanisms to run experiments and define the specific search components for each metaheuristic. Apart from the adaptation required for integration purposes, JCLEC+ enables the independence  between the foundations underlying each metaheuristic paradigm and those parts that may require some specific adaptation to deal with MOPs and MaOPs. JCLEC+ also implements a module to work with PSO, but its specification provides extension mechanisms to easily incorporate new sorts of metaheuristics in the future.

Finally, it is noteworthy that external interactions are permitted with the aim of integrating external analytical tools like datapro4j\footnote{\url{http://www.uco.es/grupos/kdis/datapro4j}}, a Java library for data processing, or R\footnote{\url{http://www.r-project.org/}}, a well-known programming language for statistical analysis. For instance, the module \textit{Reporting} within JCLEC-MO makes use of datapro4j to generate reports in different ways and formats. Additionally, R can be used to perform an in-depth analysis of either the obtained PF approximation or quality indicators by simply importing CSV files. 


\section{Software modules}\label{sec:modules}

\begin{figure*}[!t]
	\centering
		\includegraphics[width=\textwidth]{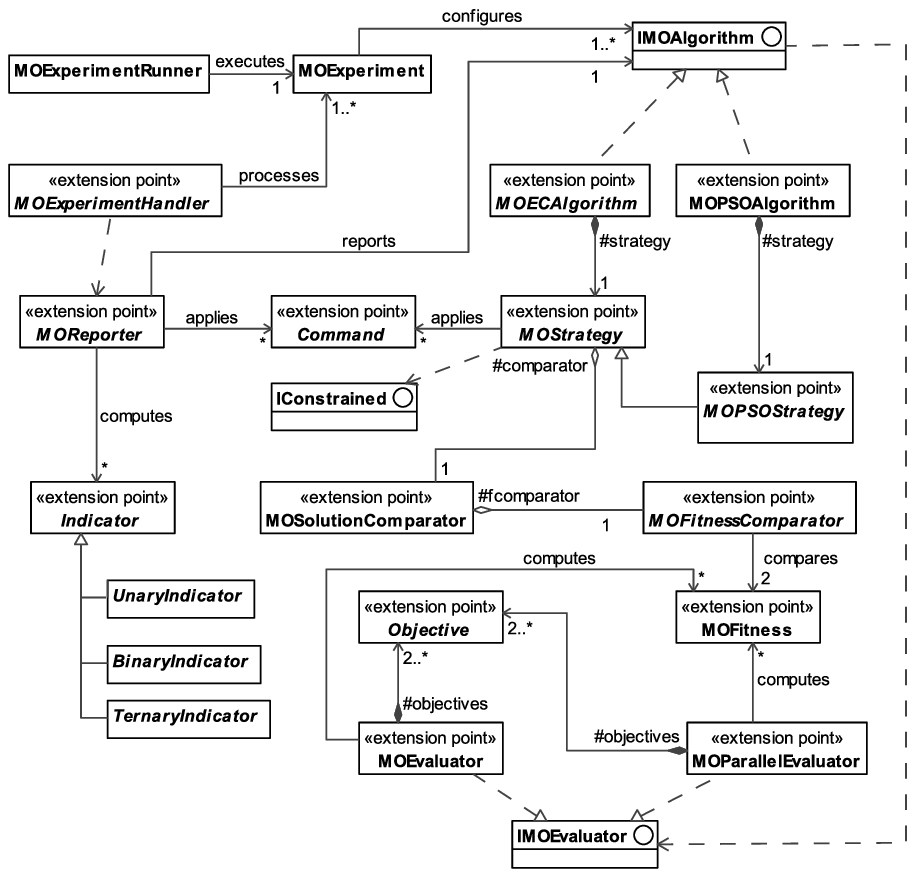}
	\caption{JCLEC-MO main classes and interfaces}
	\label{fig:core}
\end{figure*}

On the basis of the architecture explained above, this section augments the discussion on the main classes and interfaces that constitute the specification of JCLEC-MO. In more detail, Figure~\ref{fig:core} depicts the class diagram specifying the concrete design elements and their relationships. Abstract classes are written in italics and extension points are identified by the stereotype \textit{$\ll$extension point$\gg$}. Elements related to the implementation of multi-objective algorithms are explained at first, followed by those referring to the problem definition and the utilities provided for experimentation, validation and reporting.

\subsection{Algorithms}\label{subsec:algorithms}

Design extensibility and updated availability (see Table~\ref{tab:design-principles}) are two relevant design principles well formalized in the implementation of multi- and many-objective algorithms in JCLEC-MO. To this end, an explicit distinction is made between phases of the iterative process defined by a concrete metaheuristic, namely \emph{algorithm}, and the specific steps requiring some adaptation to deal with MOPs and MaOPs, referred as \emph{multi-objective strategy}. It is noteworthy that algorithms have common characteristics regarding the way in which solutions are handled. Similarly, strategies can be reused across different algorithms, and new multi-objective approaches can be incorporated by simply coding the steps as determined by the strategy.

Within the module \emph{algorithms}, the interface \emph{IMOAlgorithm} declares the common operations to be implemented by every multi-objective algorithm, independently of its metaheuristic model. There are currently two classes implementing this interface. Firstly, \emph{MOECAlgorithm} encapsulates the iterative process performed by algorithms based on evolutionary computation. This is an abstract class, so it is specialized to provide different types of evolutionary algorithms like genetic algorithms or genetic programming~\citep{Boussaid13}. Secondly, the abstract class \emph{MOPSOAlgorithm} is in charge of adapting the PSO paradigm to MOO. Other metaheuristic models can be similarly incorporated, what easily benefits extensibility and reusability of code. On the other hand, multi-objective strategies are represented by the abstract class \emph{MOStrategy}. Based on the so-called \emph{Strategy} design pattern~\citep{Gamma13}, this class specifies the steps of the MOO approach, allowing its combination with any current or future sort of algorithm. Analogously, \emph{MOPSOStrategy} is specifically devoted to include the steps followed by multi-objective PSO.

\begin{figure*}[!t]
	\begin{center}
	\subfloat[b][MOEAs]{
		\includegraphics[width=\textwidth]{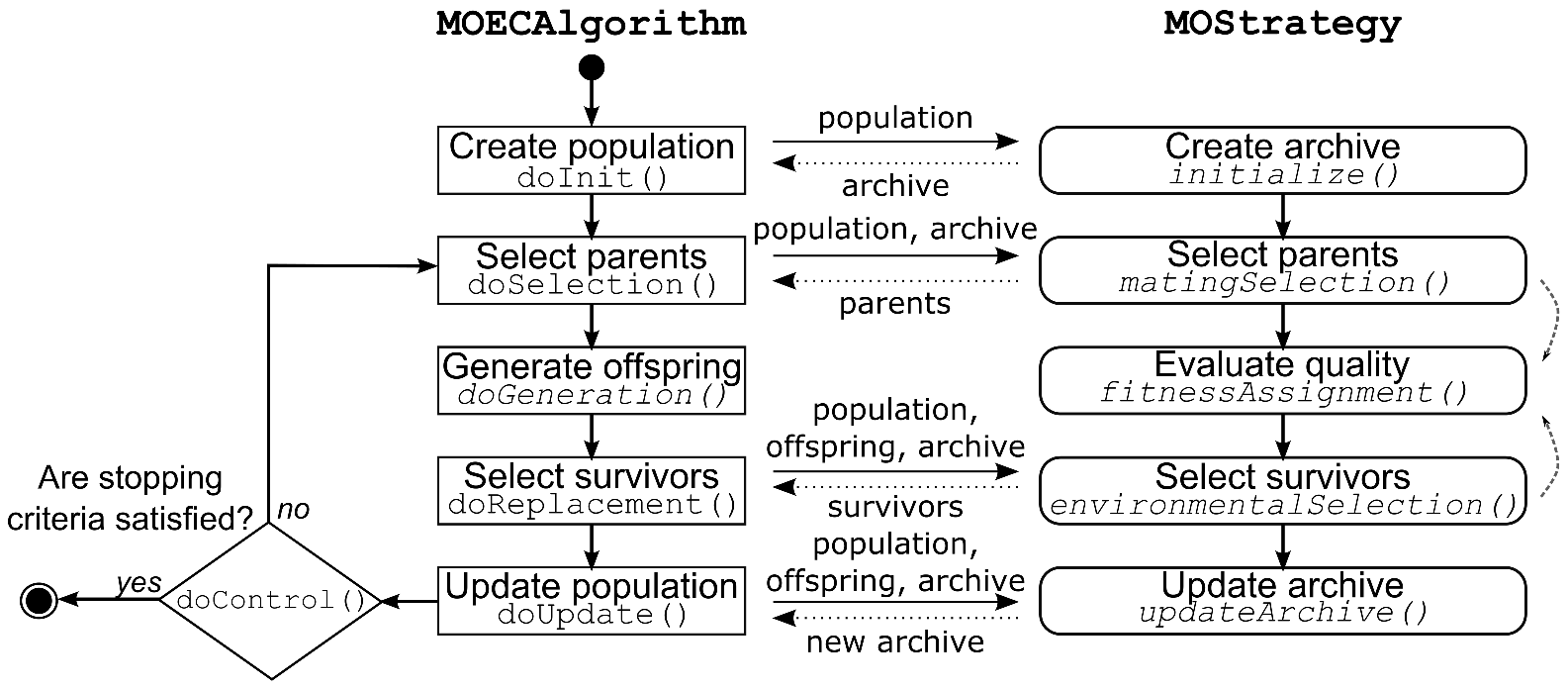}
		\label{fig:moeas}
	}
    
	\subfloat[b][MOPSO]{
		\includegraphics[width=\textwidth]{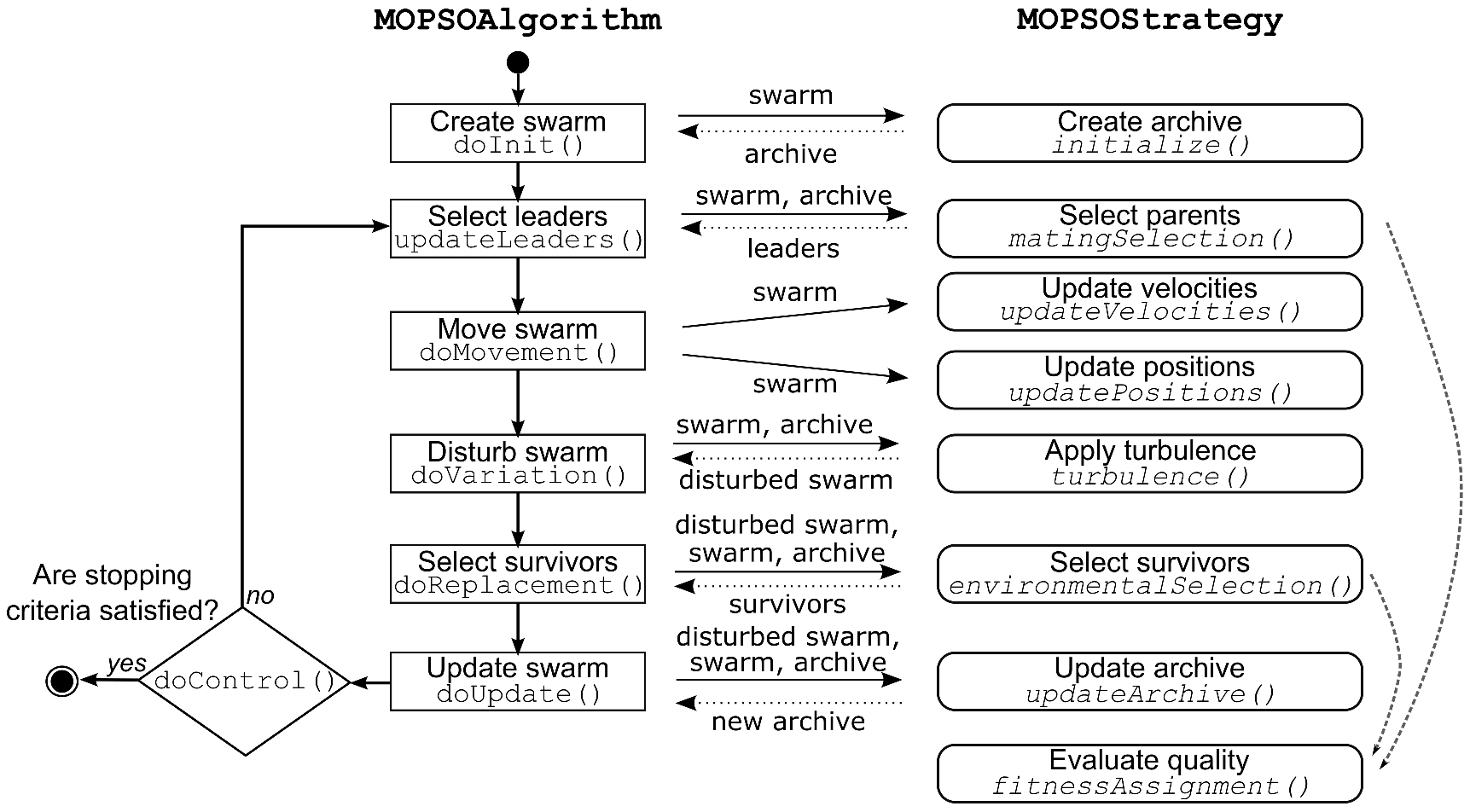}
		\label{fig:mopso}
	}
	\caption{Collaboration between algorithms and strategies}
	\label{fig:algorithms}
	\end{center}
\end{figure*}

For a more precise description, Figure~\ref{fig:algorithms} shows collaborations between algorithms and strategies, where text in italics stands for abstract methods. Interactions are depicted for both MOEAs (see Figure~\ref{fig:moeas}) and PSO (see Figure~\ref{fig:mopso}). For instance, focusing on the search conducted by MOEAs, method invocation and return is performed as follows:

\begin{enumerate}
	\item Initialization. The algorithm creates the initial population according to the selected encoding and problem characteristics. Next, the strategy is invoked to create the initial archive, if required.

	\item Parent selection. The algorithm delegates this step to the strategy, where parents are selected considering both the population and archive. To this end, the strategy invokes \emph{fitnessAssignment()} to execute its own evaluation mechanism.
    
	\item Offspring generation. The algorithm executes the genetic operators.
	
	\item Replacement. The algorithm asks the strategy to proceed with the selection of survivors, which are picked from the current population, the offspring and the archive altogether. The strategy may invoke a quality evaluation mechanism to complete this step, if required.
	
	\item Population update. Survivors constitute the new population, and the algorithm asks the strategy to update the archive.
	
	\item Stop criteria. The algorithm checks whether every stop condition is satisfied, such as the maximum number of generations or evaluations, the existence of a solution with an admissible fitness value, or any other user-defined criterion.
\end{enumerate}

Notice that a collaboration schema with a number of common steps is proposed for multi-objective PSO, in compliance with design principles, promoting code understandability and reusability. Other collaborations between algorithm and strategy in PSO respond to the particularities of multi-objective PSO proposals. For instance, the algorithm delegates the movement of particles and the execution of the variation mechanism to the strategy, as shown in Figure~\ref{fig:mopso}.

In the current version, JCLEC-MO includes an implementation for each of the 
following types of EAs: genetic algorithm (with or without mutation), evolution strategy, genetic programming and evolutionary programming.
Regarding PSO, a standard algorithm with a turbulence mechanism is implemented as a subclass of \emph{MOPSOAlgorithm}. Furthermore, the framework provides an extensive collection of representative strategies belonging to diverse families of approaches~\citep{Wagner07,Li15}. Table~\ref{tab:list-algorithms} shows the list of available strategies, including information about the sort of MOO approach and the year of publication. All these strategies are also adapted to deal with constrained problems.

\begin{table}[!t]
\begin{center}
\caption{Available multi-objective and many-objective strategies}
\begin{tabular}{|l|c|c|} \hline
\textbf{Algorithm}												&\textbf{Acronym}	&\textbf{Year}\\\hline
\multicolumn{3}{|l|}{\textit{Based on Pareto dominance}}\\\hline
Pareto Archived Evolution Strategy 								& PAES 				& 2000	\\
Strength Pareto Evolutionary Algorithm 							& SPEA2				& 2001	\\
Non-dominated Sorting Genetic Algorithm II						& NSGA-II 			& 2002	\\
Multi-Objective CHC Algorithm 									& MOCHC				& 2007	\\
\hline
\multicolumn{3}{|l|}{\textit{Based on landscape partition}}\\\hline
$\epsilon$ Multi-Objective Evolutionary Algorithm				& $\epsilon$-MOEA 	& 2002	\\
Grid-based Evolutionary Algorithm								& GrEA				& 2013	\\
\hline
\multicolumn{3}{|l|}{\textit{Based on indicators}}\\\hline
Indicator Based Evolutionary Algorithm 							& IBEA 				& 2004	\\                        
S Metric Selection Evol. Multi-Objective Algorithm				& SMS-EMOA 			& 2007	\\
Hypervolume Estimation Algorithm								& HypE 				& 2011	\\
\hline
\multicolumn{3}{|l|}{\textit{Based on decomposition}}\\\hline
Multi-Objective Evol. Alg. based on Decomposition 				& MOEA/D 			& 2007	\\
\hline
\multicolumn{3}{|l|}{\textit{Based on reference points}}\\\hline
Non-dominated Sorting Genetic Algorithm III 					& NSGA-III 			& 2014	\\
Reference Vector-guided Evolutionary Algorithm					& RVEA 				& 2016	\\
\hline
\multicolumn{3}{|l|}{\textit{Based on preferences}}\\\hline
Preference-based adaptive region of interest 					& PAR				& 2016	\\
\hline
\multicolumn{3}{|l|}{\textit{PSO algorithms}}\\\hline
Multi-Objective Particle Swarm Optimizer						& OMOPSO 			& 2005	\\
Speed-constrained Multi-objective PSO							& SMPSO 			& 2009	\\
\hline
\end{tabular}
\label{tab:list-algorithms}
\end{center}
\end{table}

\subsection{Problem-specific elements}\label{subsec:domain}

Solving real-world optimization problems may require coding problem-specific elements, as provided by modules \emph{EA} and \emph{PSO} within JCLEC+. They include classes to represent and modify solutions. Additionally, the module \emph{problem} of JCLEC-MO accepts user-provided code to evaluate their quality and specialize the solution encoding, if needed.

Notice that solutions in JCLEC-MO can be represented using all the available encodings in JCLEC~\citep{Ventura08}, including binary, integer, real or tree structures. Existing genetic operators can manage all these types of encodings, offering a great variety of combinations to be chosen. Additionally, a specialization of the real encoding enclosing velocity and memory properties to represent particles is provided by the module \emph{PSO}. All these elements serve to guarantee domain adaptability.

On the other hand, objective functions should be defined as an extension of the \emph{Objective} class. An evaluation method has to be coded in order to compute the objective value for a given solution, as well as to specify the limits of the function and indicate whether it should be minimized or maximized.

JCLEC-MO provides an explicit mechanism to include constraints as part of a generic problem definition. With this aim, the \emph{IConstrained} interface should be implemented by a class representing candidate solutions of a constrained problem. Other participants of the search process like evaluators and strategies can access the information about the solution feasibility by calling its methods \emph{isFeasible()} and \emph{degreeOfInfeasibility()}, which specify whether a solution is feasible or not and the overall degree of constraints violation, respectively.

\subsection{Experiments}\label{subsec:experiment}

The evaluation of solutions is carried out as a separate step, properly controlled by the so-called \emph{Evaluator} class. \emph{IMOEvaluator} interface within the module \emph{experiments} declares the list of operations that any evaluator should provide to an algorithm when solving either a MOP or a MaOP. JCLEC-MO provides two classes aimed at evaluating solutions either sequentially (\emph{MOEvaluator}) or in parallel (\emph{MOParallelEvaluator}), depending on preferences.
These classes handle the set of objective functions implemented for the specific problem, iteratively invoking them in order to get all the objective values for a given solution. 

Regardless of the type of evaluator, objective values are always encapsulated within \emph{MOFitness} class, which conveniently allows assigning a single quality value to a solution, as some multi-objective algorithms like SPEA2 and IBEA do~\citep{Zhou11}. In contrast, other algorithms compute additional properties to make further comparisons, such as the crowding distance in NSGA-II~\citep{Coello07}. JCLEC-MO defines subclasses that extend \emph{MOFitness} to allow these properties to be part of the fitness object. According to the \emph{Prototype} design pattern, the evaluator will create fitness objects from one given prototypical object defined by the strategy. This process guarantees the independence of the different search components and enhances extensibility of the existing algorithms.

Once all the components of an algorithm have been defined or selected among those available, JCLEC-MO will proceed with its configuration and execution. Notice that search algorithms are comprised of several elements that might require parameters, which should be properly configured and tuned. In addition, since randomness is present in these algorithms, good practices recommend accurately assessing their performance by executing them several times and then aggregating the obtained results. JCLEC-MO includes diverse tools aimed at facilitating this task and putting all the pieces together, what is commonly referred as an experiment. Within the \emph{experiments} module, an experiment is represented by the \emph{MOExperiment} class. It aggregates a set of algorithm configurations, i.e. components and their parametrization, to be deployed and executed by the \emph{MOExperimentRunner} class. Then, a sequence of post-processing steps could be specified to perform a detailed analysis considering one of more experiments. Following the precepts of the \emph{Chain of responsibility} design pattern~\citep{Gamma13}, each individual post-processing task is defined as a separate class that extends a general handler named \emph{MOExperimentHandler}. These handlers are then connected to determine the ordered sequence of steps required by the post-processing flow, depending on each specific case. The generation of R plots and the application of statistical tests are examples of the functionalities currently available. In the latter case, JCLEC-MO handlers can make use of the wrappers available in datapro4j to execute either parametric, e.g. Anova or t-test, or non-parametric tests, e.g. Wilcoxon, Friedman or Kruskal-Wallis. On the other hand, the engineer could also define his/her own analytical procedure as a R script to be directly executed from the framework.

\subsection{Quality indicators and reporters}\label{subsec:indicators}

The \emph{indicators} module includes the abstract definition of a performance measure (\emph{Indicator} class), which can be specialized into unary, binary and ternary indicators, depending on the number of PFs in which they operate. The complete list of indicators is shown in Table~\ref{tab:list-indicators}. Quality indicators can be reported at a user-defined frequency during or after the search. In addition, the \emph{reporting} module provides a collection of classes to generate different types of reports, they all inherited from \emph{MOReporter} class. It is worth observing that reporters also provide access to the set of non-dominated solutions and the corresponding PF.

\begin{table}[!t]
\begin{center}
\caption{Available quality indicators}
\scalebox{0.95}{
\begin{tabular}{|ll|} \hline
\multicolumn{2}{|l|}{\textit{Unary indicators}}\\\hline	
\multicolumn{2}{|l|}{Overall non-dominated vector generation (ONVG)}\\
Hypervolume ($HV$) & Spacing\\\hline
\multicolumn{2}{|l|}{\textit{Binary indicators}}	\\\hline
Epsilon ($I_{\epsilon}$) 			& Additive epsilon ($I_{\epsilon+}$)\\
Spread ($S$) 						& Generalized Spread ($\Delta S$)\\
Generational distance ($GD$)		& Inverted generational distance ($IGD$)\\
Error ratio ($ER$) 					& Maximum PF error ($ME$)\\
Hyperarea ratio ($HR$) 				& Two set coverage\\
$R2$ 								& $R3$\\
Non-dominated vector addition ($NVA$) & $ONVG$ ratio\\\hline
\multicolumn{2}{|l|}{\textit{Ternary indicators}}\\\hline
\multicolumn{2}{|l|}{Relative progress}\\
\hline
\end{tabular}
}
\label{tab:list-indicators}
\end{center}
\end{table}

\subsection{Utilities}\label{subsec:utilities}

JCLEC-MO provides additional types of low-level \emph{utilities} and constructors that enable saving development time and reducing the required effort:

\begin{itemize}
	\item \emph{Commands} are recurrent operations like objective transformations and sorting methods involving the whole population. Based on the so-called \emph{Command} design pattern~\citep{Gamma13}, they are properly implemented as highly configurable classes that could be called from strategies and reporters.
	
	\item Along the search process, strategies may need to perform comparisons between solutions. With this aim, two different categories of \emph{comparators} are implemented: \emph{MOFitnessComparator} receives two fitness objects to make the comparison, whereas \emph{MOSolutionComparator} considers complete solutions by setting a primary criterion (e.g., feasibility) to decide which solution would be preferred. If both solutions are equivalent, the result would depend on the comparator at the fitness level.
    
	\item \emph{Diversity preservation mechanisms} are frequently based on computing distances between solutions within the objective space. JCLEC-MO provides different implementations for \emph{IDistance} interface, adopted from JCLEC, that compute both the euclidean and Manhattan distances.

	\item MOFs usually provide \emph{benchmarks} to facilitate the comparison of new proposals against the state-of-the-art. They can also be used to run and test algorithms or as code templates for addressing new MOPs, which can be helpful for less experienced users of the suite. At least one benchmark is provided per type of encoding: the knapsack problem (binary), the traveling salesman problem (integer), the DTLZ and ZDT families (real)~\citep{Coello07}, and a symbolic regression problem (tree). The availability of benchmarks could be easily extended too.
\end{itemize}

\section{An illustrative running example}\label{sec:example}

JCLEC-MO enables engineers to integrate optimization approaches in their industrial applications in just a few simple steps. In this section, the resolution of a water resource management problem is presented as a case study. It is addressed from a many-objective perspective, and its representation, coding and configuration is explained below, as well as the way in which outcomes are subsequently processed and analyzed.

\subsection{Problem representation}\label{subsec:problem}

The Water Resource Management (WRM) problem~\citep{Ray01} consists in finding the optimal planning for a drainage system. This real-world problem, for which the performance of many-objective algorithms has been already reported~\citep{Deb14,Asafuddoula15}, is defined in terms of three decision variables: local detention storage capacity ($x_1 \in [0.01,0.45]$), maximum treatment rate ($x_2 \in [0.01,0.10]$) and maximum allowable overflow rate ($x_3 \in [0.01,0.10]$). Five objective functions, which are conceived to be minimized, compute the following aspects: drainage network cost ($f_1$), storage facility cost ($f_2$), treatment facility cost ($f_3$), expected flood damage cost ($f_4$) and expected economic loss due to flood ($f_5$). They are properly defined as follows:

\begin{align}
	&f_1(x) = 106780.37\cdot(x_2+x_3)+61704.67\\
	&f_2(x) = 3000.00\cdot x_1\\
	&f_3(x) = 30570.00\cdot 0.02289\cdot x_2/(0.06\cdot2289.0)^{0.65}\\
	&f_4(x) = 250.00\cdot 2289.00\cdot e^{-39.75\cdot x_2+9.90\cdot x_3+2.74}\\
	&f_5(x) = 25.00\cdot (1.39/(x_1\cdot x_2))+4940.0\cdot x_3 + 2.74
\end{align}

In addition, the WRM problem presents the following seven constraints
(a more detailed description of the problem can be found in~\citep{Ray01}):

\begin{align}
	&g_1(x) = 0.00139/(x_1\cdot x_2) + 4.94\cdot x_3 - 0.08 \le 1\\
	&g_2(x) = 0.000306/(x_1\cdot x_2) + 1.082\cdot x_3 - 0.0986 \le 1\\
	&g_3(x) = 12.307/(x_1\cdot x_2) + 49408.24\cdot x_3 + 4051.02 \le 50000\\
	&g_4(x) = 2.098/(x_1\cdot x_2) + 8046.33\cdot x_3 - 696.71 \le 16000\\
	&g_5(x) = 2.138/(x_1\cdot x_2) + 7883.39\cdot x_3 - 705.04 \le 10000\\
    &g_6(x) = 0.417/(x_1\cdot x_2) + 1721.26\cdot x_3 - 136.54 \le 2000\\
	&g_7(x) = 0.164/(x_1\cdot x_2) + 631.13\cdot x_3 - 54.48 \le 550
\end{align}

\subsection{Translating the problem into code in JCLEC-MO}\label{subsec:implem}

Firstly, JCLEC-MO requires some problem-specific elements to be defined and implemented: $(a)$ the type of encoding representing candidate solutions, $(b)$ the objective functions and $(c)$ the evaluator of solutions. Because of the highly modular design of the framework, all these elements should be implemented only once, since they can be easily combined with the rest of classes of JCLEC-MO if other different algorithms want to be tested. Thus, engineers can find a competitive algorithm with just a few lines of codes and a lightweight configuration process.

For the problem being addressed, a real encoding has been selected, which perfectly allows the application of both EAs and PSO and opens up the possibility of choosing from a variety of compatible genetic operators provided by the framework. Additionally, the WRM problem presents several constraints to be considered. As explained in Section~\ref{sec:design}, JCLEC+ provides the base implementation of a PSO solution, named \emph{Particle}, which is actually a specialization of \emph{RealArrayIndividual} class. In addition, since it is a constrained problem, this class should implement \emph{IConstrained} interface, as illustrated in Listing~\ref{codeSolution}.

\begin{lstlisting}[language=Java,caption={Solution encoding for the WRM problem},captionpos=b,label=codeSolution]
public class WRMSolution extends Particle implements IConstrained {
	boolean isFeasible; 												/** The solution is feasible */
	double degreeOfInfeasibility;			/** The degree of infeasibility */
	...
}
\end{lstlisting}

Next, objective functions are developed as a specialization of \emph{Objective}. As an example, the code required to implement $f_1$ is shown in Listing~\ref{codeF1}, where $x_2$ and $x_3$ are extracted from the array encoding decision variables, i.e. the genotype for EAs or the particle position in PSO (lines 3-5).Then, the current value for the function (line 6) is computed, and the fitness object returned (lines 7-8).

\begin{lstlisting}[language=Java,caption={Evaluation of the first objective of the WRM problem},captionpos=b,label=codeF1]
public class F1 extends Objective {
	public IFitness evaluate(IIndividual solution) {
		double [] genotype = ((RealArrayIndividual)solution).getGenotype();
		double x2 = genotype[1];
		double x3 = genotype[2];
		double objectiveValue = 106780.37 * (x2 + x3) + 61704.67;
		IFitness fitness = new SimpleValueFitness(objectiveValue);
		return fitness;
	}
}
\end{lstlisting}

As part of the evaluation process, a subclass of \emph{MOEvaluator} checks whether the constraints are met. For illustrative purposes here, the evaluator only accumulates the degree of violation of each constraint. As shown in Listing~\ref{codeEvaluator}, all constraints are checked (lines 5-13) right after evaluating all objectives, what is actually performed by invoking the \emph{super} object (line 4). Finally, the decision about the solution feasibility is made (lines 14-19).

\begin{lstlisting}[language=Java,caption={Evaluator for the WRM problem},captionpos=b,label=codeEvaluator]
public class WRMEvaluator extends MOEvaluator {
	protected void evaluate(IIndividual solution) {	
		// Call super implementation (evaluate objective functions)
  super.evaluate(solution);		
		// Check constraints
  double [] genotype = ((RealArrayIndividual)solution).getGenotype();
  double x1 = genotype[0], x2 = genotype[1], x3 = genotype[2];
		double [] g = new double[7];
		g[0] = (0.00139/(x1*x2)+4.94*x3-0.08) - 1; 						// g1 constraint function
		...																                              // Rest of constraints
		double total = 0.0;
		for(int i=0; i<7; i++) 	                         // Compute the degree of constraint violation
			  if(g[i] > 0) total += g[i];
		if(total > 0){			                                // Infeasible solution
     ((IConstrained)solution).setFeasible(false);
     ((IConstrained)solution).setDegreeOfInfeasibility(total);
   } else 					                                    // Feasible solution
	    ((IConstrained)solution).setFeasible(true);
}
\end{lstlisting}

\subsection{Configuration and execution}\label{subsec:conf}

In real-world applications, it may be extremely difficult for the engineer to know beforehand which algorithm is the most appropriate to address a given problem like WRM. In fact, it is likely that engineers might not be confident about neither the metaheuristic that best suits the problem nor its specific parametrization. In this context, given that the suite provides flexible and highly-customized implementations, conducting a comparison of the performance of different alternatives becomes easy and convenient. For the WRM problem under study, four algorithms will be applied and compared, that is, three many-objective evolutionary algorithms (GrEA, HypE and NSGA-III) and one PSO approach (SMPSO). Although the latter was originally proposed as a multi-objective approach, it is founded on the $\epsilon$-dominance principle, which is frequently used in many-objective optimization.

Following the JCLEC philosophy, in JCLEC-MO each algorithm is set up in terms of a XML-based configuration file (see Listing~\ref{confFile}). It contains values of both general parameters, such as the population size and the number of generations (lines 14-15), and other specific elements requiring the injection of external code, such as the aforementioned objective functions (lines 7-10). Notice that the metaheuristic algorithm (line 2) is independent of the multi-objective strategy (line 3), whose parameters should be also configured (line 4). Genetic operators can be chosen from those available in JCLEC~\citep{Ventura08}. In this example, BLX-$\alpha$ crossover (line 12) and polynomial mutation (line 13) have been chosen, and probabilities are configured too. Next, a list of random seeds is provided to perform independent runs of the method (lines 16-19). Finally, a number of reporters can be configured to save outcomes for further processing. For this experiment, two different reporters are selected: \emph{MOParetoFrontReporter} is provided to store the PF (lines 20-22), and \emph{MOComparisonReporter} serves to compute two unary indicators (lines 23-31), namely hypervolume and spacing.

\begin{lstlisting}[language=XML,caption={Example of a configuration file},captionpos=b,label=confFile]
<experiment>
	<process algorithm-type=``net.sf.jclec.mo.algorithm.MOGeneticAlgorithm">
		<mo-strategy type=``net.sf.jclec.mo.strategy.constrained.ConstrainedHypE" >
			<sampling-size>10000</sampling-size>
		</mo-strategy>
  <evaluator type=``net.sf.jclec.mo.problem.wrm.WRMEvaluator">
			<objectives>	 
				<objective type=``net.sf.jclec.mo.problem.wrm.F1" maximize=``false"/>
				...
			</objectives>
		</evaluator>
  <recombinator type=``net.sf.jclec.realarray.rec.BLXAlphaCrossover2x2" rec-prob=``0.9" />
	 <mutator type=``net.sf.jclec.realarray.mut.PolynomialMutator" mut-prob=``0.15" />
  <population-size>100</population-size>
  <max-of-generations>500</max-of-generations>
  <rand-gen-factory multi=``true">
	   <rand-gen-factory type=``net.sf.jclec.util.random.RanecuFactory" seed=``123456789"/>
       <rand-gen-factory type=``net.sf.jclec.util.random.RanecuFactory" seed=``234567891"/>
       ...
	 </rand-gen-factory>
  <listener type=``net.sf.jclec.mo.listener.MOParetoFrontReporter">
	   <report-title>WRMExperiment</report-title>
  </listener>
  <listener type="net.sf.jclec.mo.listener.MOComparisonReporter">
			<report-title>WRMExperiment</report-title>
			<number-of-algorithms>4</number-of-algorithms>
			<number-of-executions>10</number-of-executions>
			<indicators>
				<indicator type="net.sf.jclec.mo.indicator.Hypervolume"/>
				<indicator type="net.sf.jclec.mo.indicator.Spacing"/>
			</indicators>
  </listener>
	</process>
</experiment>
\end{lstlisting}

One XML configuration file could be deployed per algorithm. Then, the MOO experiment is invoked by a very simple Java program, as illustrated in Listing~\ref{codeMain1}. Firstly, configuration files are processed (lines 4-5) and, secondly, the experiment is run (lines 7-10).

\begin{lstlisting}[language=Java, caption={Creation of the experiment for the case study},captionpos=b,label=codeMain1]
public class WRMCaseStudy {
	public static void main(String [] args){
		// Declaring all the experiments
		MOExperiment experiment = new MOExperiment();
		experiment.addConfigurationsFromDirectory("./configuration-files/");
		// Running experiments
		MOExperimentRunner runner = new MOExperimentRunner();
		for(int i=0; i<experiment.getNumberOfConfigurations(); i++){
			runner.executeSequentially(experiment.getConfiguration(i));
		}
		...
}
\end{lstlisting}

\subsection{Post-processing and analysis of results}\label{subsec:post}

Depending on the configuration of reporters, different types of outcomes will be saved. In this case, the output directory will contain the PF found by each algorithm, as well as the values of the selected unary indicators, i.e. hypervolume and spacing. Once all the algorithms have been executed, a chain of responsibility can be constructed to specify the post-processing procedure. It is worth noting that there is not a standard procedure to analyze the performance of many-objective algorithms. Nevertheless, experimental studies often report a set of quality indicators or the range of values for the objectives~\citep{Lucken14}. These results can be supported by graphics representing the obtained PFs~\citep{Walker13}. For the WRM case study, we want to generate some plots to visually analyze the obtained PFs and compute the value of several binary indicators, which require a reference PF. The latter will be constructed in terms of all the non-dominated solutions found by the algorithms. More in detail, the chain of responsibility for this case study is organized according to the following ordered steps:

\begin{enumerate}
	\item Obtaining one single PF for each algorithm considering all its executions.
	\item Creating a reference PF taking the previous PFs as a basis.
	\item Scaling the values of the PFs generated from steps 1 and 2 for an easier interpretation and fair comparison of results.
	\item Generating boxplots for unary quality indicators.
	\item Creating parallel coordinates plots to visualize the PFs.
	\item Computing binary quality indicators using the reference PF, e.g. $I_\epsilon$, $I_{\epsilon +}$, $\Delta S$, $GD$, $IGD$ and $ME$.
    \item Applying the Kruskal-Wallis statistical test to reveal whether there are significant differences in terms of the hypervolume and spacing indicators.
\end{enumerate}

Steps from 1 to 6 are already available in JCLEC-MO, so they only need to be invoked. As can be seen in Listing~\ref{codeMain2}, handlers require a boolean array specifying whether each objective function should be maximized ($true$) or minimized ($false$), as well as the path to the reporting directory (lines 6-7). The chain is then constructed by indicating which handler should be executed in the following step by means of \emph{setSuccessor()} method (lines 14-15). Once the chain is completely declared, it could be executed by simply calling the first handler (line 18).

\begin{lstlisting}[language=Java, caption={Post-processing with handlers for the case study},captionpos=b,label=codeMain2]
public class WRMCaseStudy {
	public static void main(String [] args){
		// JCLEC runner
		...
  // Post-processing outputs
  boolean [] objs = new boolean[]{false,false,false,false,false};
  String reportDirectory = "./reports/WRMExperiment";
  ...
  // Create handlers
		MOExperimentHandler handler1 = new GenerateAlgorithmPF(reportDirectory,objs);
		MOExperimentHandler handler2 = new GenerateReferencePF(reportDirectory,objs);
   ...
		// Create chain of handlers
		handler1.setSuccessor(handler2);
		handler2.setSuccessor(handler3);
		...
		// Execute the complete process
		handler1.process();
}
\end{lstlisting}

A new handler has been implemented for the execution of the Kruskal-Wallis test (step 7). This test will serve to prove the hypothesis that the four algorithms perform similarly for a particular unary indicator. As can be observed in Listing~\ref{codeTest}, the handler firstly loads the results for the hypervolume and spacing indicators (line 3). Then, the test is computed for each indicator (lines 13-18) by invoking the wrapper of the R implementation available by datapro4j (RKruskal). Test outcomes, which indicate whether the hypothesis would be rejected and the level of confidence in the decision, are then processed (lines 21-22) and stored in a text file (line 5).

\begin{lstlisting}[language=Java, caption={Handler to execute the Kruskal-Wallis test},captionpos=b,label=codeTest]
public class WRMKruskalWallisTestHandler extends MOExperimentHandler {
	public void process() {
		readDatasets();
		computeTests();
		saveResults();
		if(nextHandler()!=null)
			nextHandler().process();
	}
    
 protected void computeTests() {
		RKruskal algorithm;
		this.testResults = new ArrayList<Map<String,Object>>();
		for(int i=0; i<this.indicatorResults.size(); i++) {
			// Execute the test for each indicator
			algorithm = new RKruskal(this.indicatorResults.get(i),"myData",false);
			algorithm.initialize();
			algorithm.execute();
			algorithm.postexec();

			// Get the result
			Map<String,Object> testMapResult = (Map<String,Object>)algorithm.getResult();
			this.testResults.add(testMapResult);
		}
	}
 ...
}
\end{lstlisting}

\begin{figure*}[!t]
\begin{center}
	\subfloat[][GrEA]{
		\includegraphics[width=0.5\textwidth]{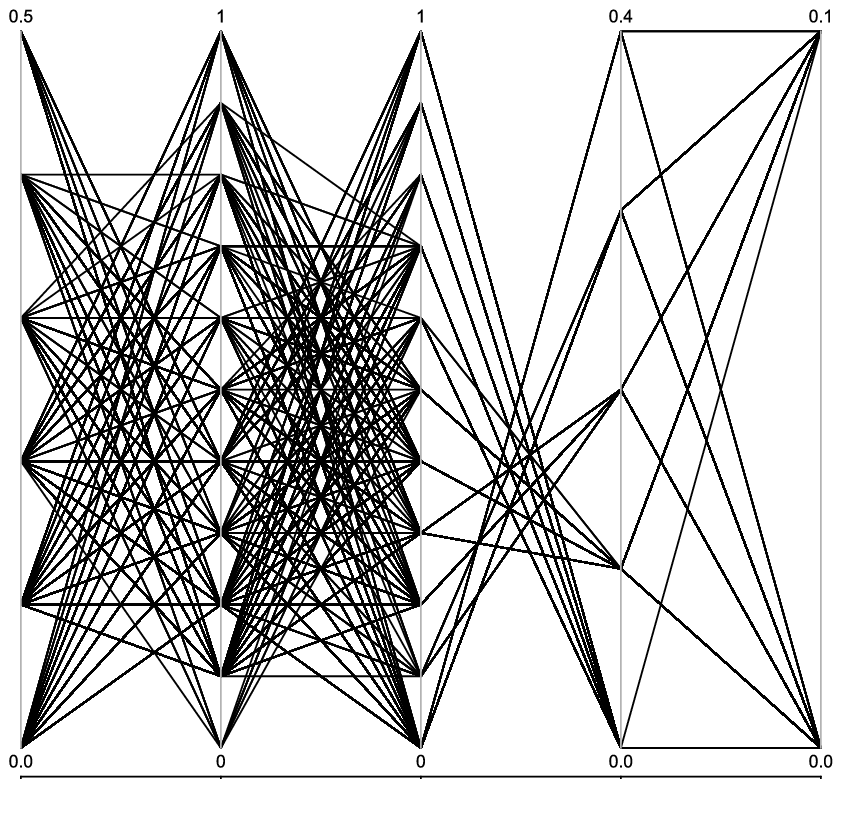}
		\label{fig:grea}
	}
	\subfloat[][HypE]{
		\includegraphics[width=0.5\textwidth]{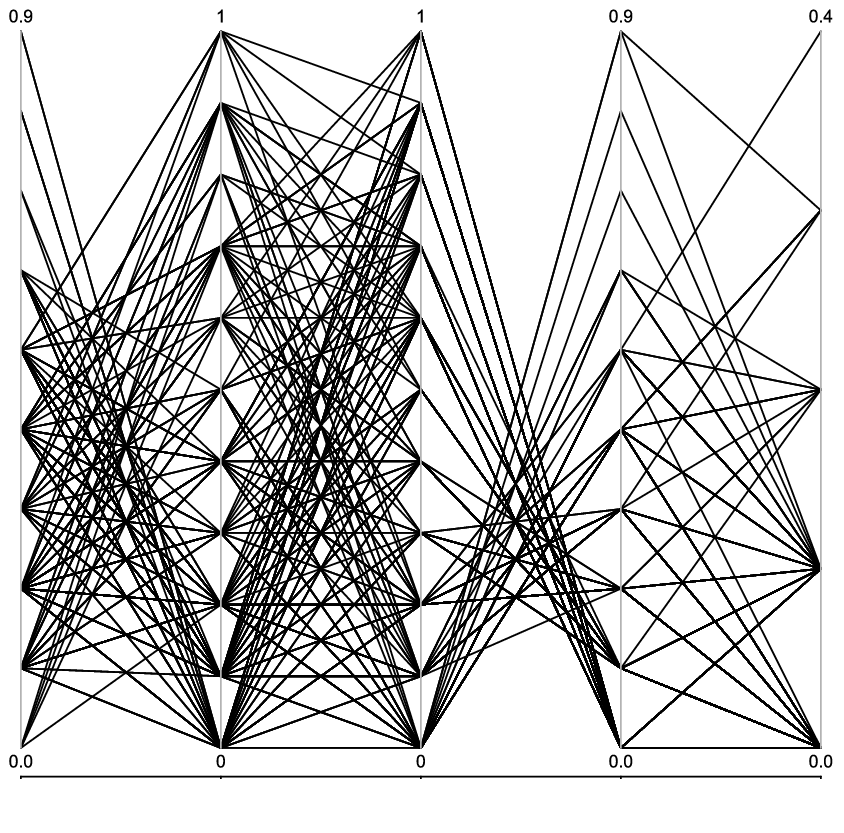}
		\label{fig:hype}
	}
    \vspace{0.01cm}
	\subfloat[][NSGA-III]{
		\includegraphics[width=0.5\textwidth]{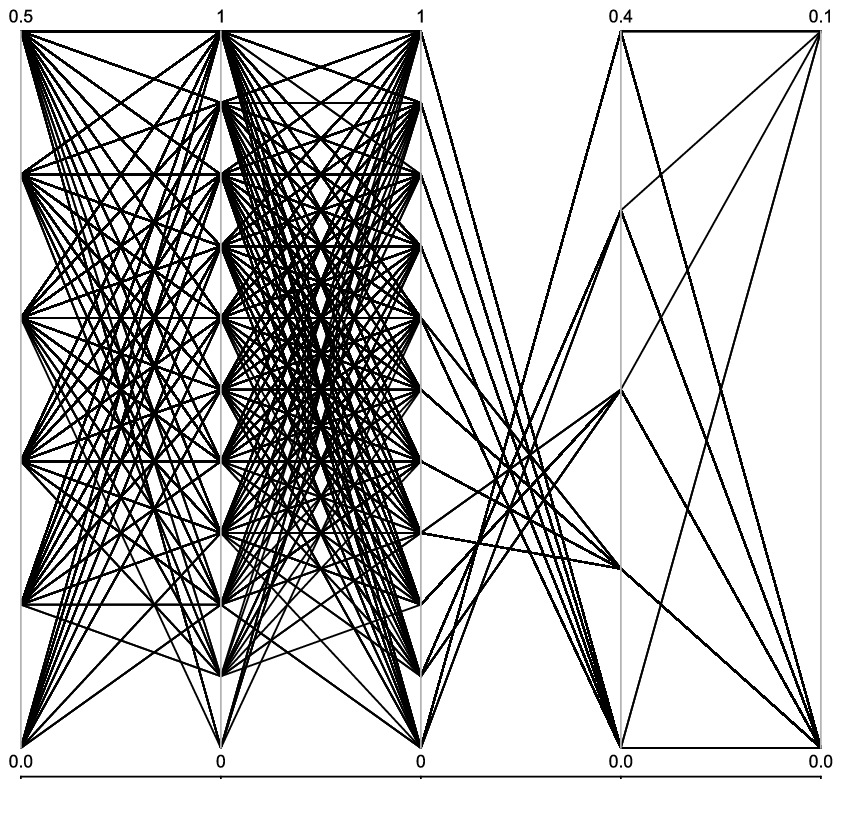}
		\label{fig:nsga3}
	}
	\subfloat[][SMPSO]{
		\includegraphics[width=0.5\textwidth]{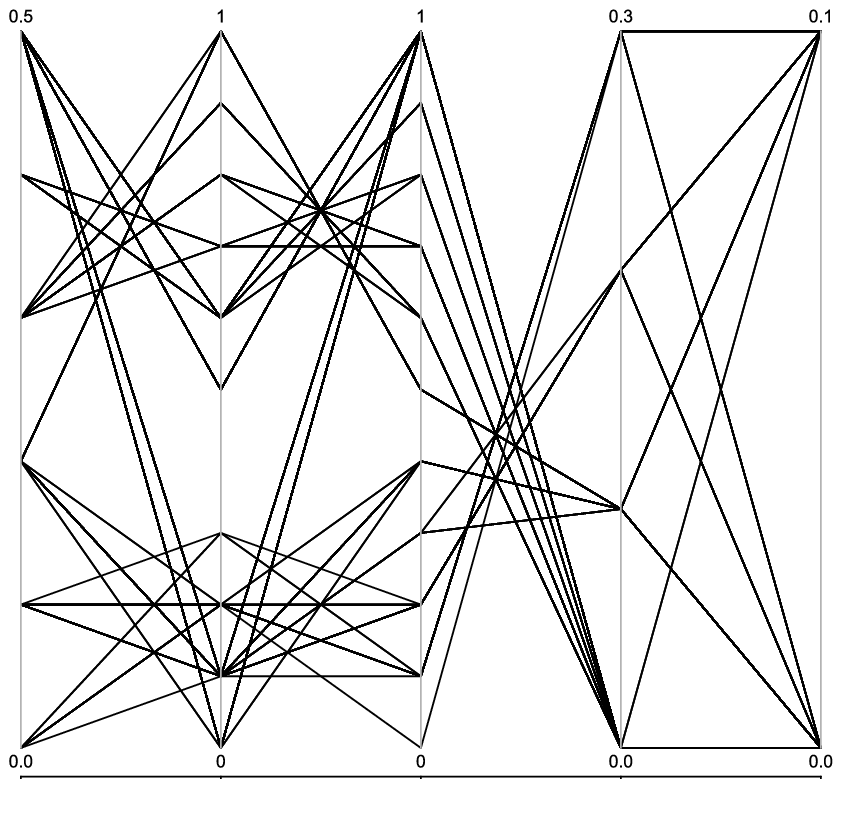}
		\label{fig:smpso}
	}
	\caption{Pareto front found by each algorithm}
	\label{fig:fronts}
\end{center}
\end{figure*}

Figure~\ref{fig:fronts} shows the parallel coordinates R plots generated by step 5, where each line represents the objective values of a non-dominated solution. These graphics are frequently used when solving MaOPs since they make the analysis of highly-dimensional PFs simpler~\citep{Walker13}. It is worth noting how each axis represents a different objective function. Consequently, a high density of lines around certain values indicates that the algorithm did not found diverse solutions. Each line representing a different solution indicates the achieved trade-off among objectives. As can be observed, the alternation of extreme values for some axes suggests the existence of conflicts between the corresponding objectives. For instance, GrEA, HypE and NSGA-III are seemingly able to find solutions having a broad range of values for $f_1$, $f_2$ and $f_3$, whereas SMPSO returns a smaller number of solutions with extreme values.

Table~\ref{tab:indicators} shows the results returned by binary quality indicators (step 6 of the post-processing chain), where the best value is written in bold. Notice that the reference PF was constructed with those PFs provided by the algorithms and all measures were minimized. As can be observed, HypE provides the best distributed PF ($\Delta S$) for the WRM problem, whilst NSGA-III returns the closest PF to the reference PF ($GD$ and $IGD$). Even though SMPSO found a smaller number of non-dominated solutions, most of them seem to belong to the reference PF, as reflected by the good values returned for $I_{\epsilon+}$ and $ME$.

\begin{figure*}[!t]
	\begin{center}
	\subfloat[][Hypervolume]{
		\includegraphics[width=0.5\textwidth]{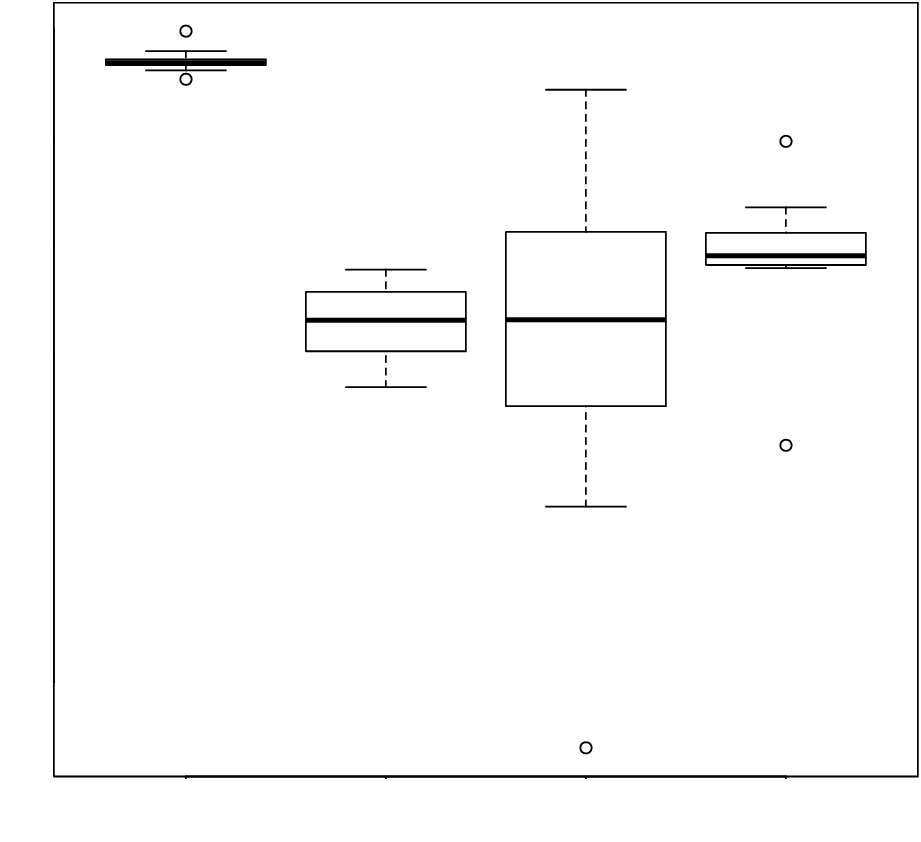}
		\label{fig:hv}
	}
	\subfloat[][Spacing]{
		\includegraphics[width=0.5\textwidth]{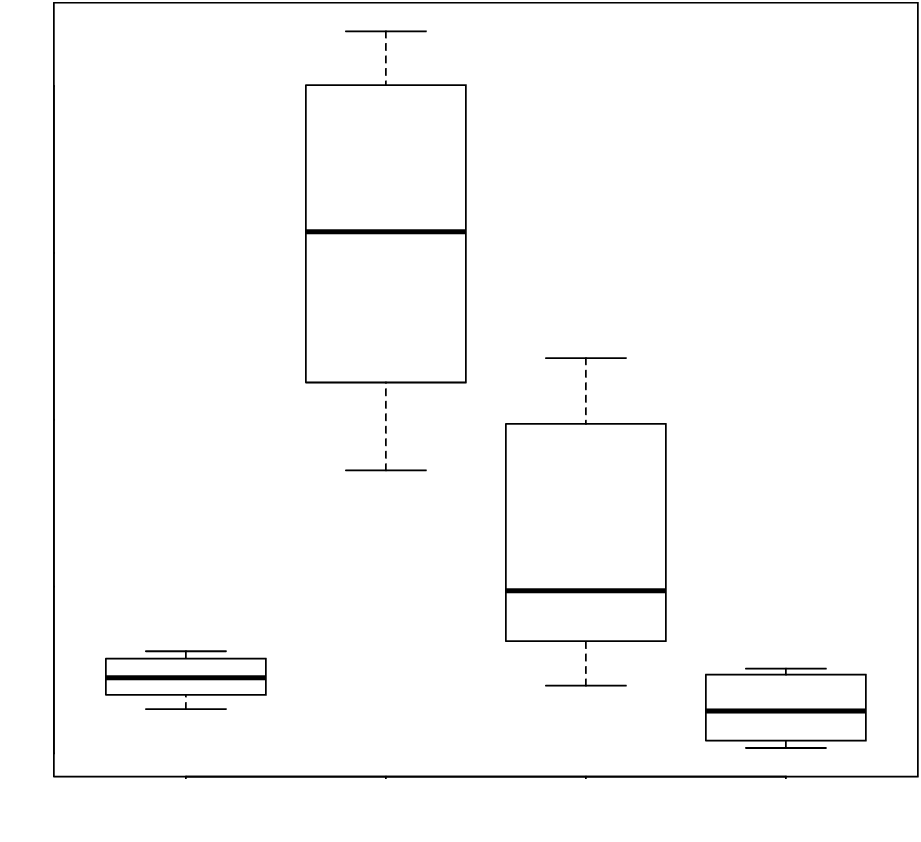}
		\label{fig:spacing}
	}
	\caption{Boxplots showing the distribution of unary indicators}
	\label{fig:indicators}
	\end{center}
\end{figure*}

Finally, since unary indicators are computed during the execution of each algorithm, we may evaluate to what extent the stochastic nature has influenced the returned values. Hence, boxplots serve to visualize the distribution of hypervolume and spacing, where higher values are preferred (see Figure~\ref{fig:indicators}). These values can also be analysed by conducting the Kruskal-Wallis test to confirm whether the observed differences are statistically significant. The obtained \emph{p-values} are 1.34E-5 for hypervolume and 4.35E-7 for spacing. In both cases, these values are clearly inferior to the usual confidence levels ($\alpha$=0.01 or $\alpha$=0.05), meaning that statistical differences among the algorithms exist.

\begin{table}[!t]
\begin{center}
\caption{Results for binary quality indicators}
\begin{tabular}{|l|c|c|c|c|c|} \hline
\textbf{Indicator}		&	\textbf{GrEA}	&	\textbf{HypE}	&	\textbf{NSGA-III}	& \textbf{SMPSO}\\\hline
$I_{\epsilon}$			&1.666049	&1.675537	&\textbf{1.382379}	&1.675127\\
$I_{\epsilon+}$			&0.049350	&\textbf{0.049149}	&0.049434	&\textbf{0.049149}\\
$\Delta S$				&0.830827	&\textbf{0.370173}	&0.674743	&0.971610\\
GD						&0.002104	&0.005367	&\textbf{0.001797}	&0.003892\\
IGD						&0.005193	&0.016245	&\textbf{0.003695}	&0.054198\\
ME						&0.097496	&0.220674	&0.090818	&\textbf{0.082821}\\
\hline
\end{tabular}
\label{tab:indicators}
\end{center}
\end{table}

\section{Differentiating characteristics of JCLEC-MO}\label{sec:comparison}

JCLEC-MO has been thoroughly developed according to the design rationale explained in Section~\ref{sec:design}, and following coding and design best practices. On this basis, this section reviews the benefits that such design criteria may offer to practitioners from an industrial environment, and compares this suite against other available MOFs (see Section~\ref{sec:related}) in relation to these aspects.

\emph{Generality preservation}.
Engineers interested in MOO, as most of non-expert users in metaheuristics, will presumably tend to make a more effective use of predefined, standard configurations of some available algorithms, investing their time in their customization and the adaptation of problem-specific methods. However, providing full support in this scenario is only possible for those MOFs providing a wide range of general, modular and pluggable components, easy to combine and parametrize~\citep{Gagne06}. Most current state-of-the-art MOFs satisfy generality preservation in terms of both problem-specific components, i.e. encodings and operators, and technique-oriented elements, i.e. algorithms and evaluators. In addition, JCLEC-MO follows this idea for MOO-specific elements too, such as fitness objects, comparators or the novel concept of strategy. In this aspect, this suite is closer to general-purpose MOFs like ECJ, EvA and Opt4j, since it takes advantage of its integrability to JCLEC in order to promote reuse of other components like genetic operators. As a result, it provides a wider range of these components than MOO-specific suites like jMetal and MOEA Framework.
    
\emph{Design extensibility}.
The increasing cooperation between academia and industry enables the development of ad-hoc solutions to address engineering-specific optimization problems. Most of them apply popular metaheuristics, and the use of novel, potentially more fitted approaches is still limited. This is where MOFs could make a relevant contribution. Within the field of MOO, some authors~\citep{Zavala13} already highlighted the importance of code extensibility and reuse. Similarly to ParadisEO, ECJ, EvA and Opt4J, JCLEC-MO maintains independence between the metaheuristic paradigm and its adaptation to MOO in favor of the aforementioned characteristics.
Furthermore, JCLEC-MO applies best practices in design, such as the use of design patterns and the explicit definition of extension points. Both aspects make future developments easier. The two metaheuristics currently available in JCLEC-MO, EAs and PSO, are seemingly the preferred paradigms among engineers~\citep{Zavala13,Kulkarni15}, probably because of historical reasons and the suitability of PSO to continuous search spaces. 

\emph{Updated availability}.
Many-objective optimization is still a developing research area. New algorithms, indicators, validation techniques, applications and studies are appearing. JCLEC-MO already provides the most extensive catalog of algorithms among general-purpose frameworks, being the only suite including algorithms specifically conceived for many-objective optimization. Only jMetal and MOEA Framework provide an implementation for some many-objective algorithms. Considering both multi- and many-objective algorithms, these two MOFs provide more algorithms than JCLEC-MO, which is normal considering that both projects have been actively developed in the last years.

\emph{Independence of the problem definition}.
JCLEC-MO does not make any assumption regarding the problem formulation, including objective functions, e.g. if they should be maximized or minimized, the representation of individuals, or the treatment of constraints. This is particularly well-suited for real-world applications, as it may increase the independence of the suite to the specific problem. Therefore, most general-purpose frameworks (ECJ, HeuristicLab, Opt4J and PaGMO) allow formulating different fashions of objective functions, which, however, is not so recurrent among MOO-specific tools: only ParadisEO-MOEO was built considering this aspect. Besides JCLEC-MO, MOEA Framework, jMetal, ECJ, EvA, Opt4J and PaGMO have some sort of support for handling constraints, which is a determining factor in the case of engineering problems~\citep{Singh16}.
    
\emph{Domain adaptability}.
Values in engineering problems are often continuous, which implies that candidate solutions could be precisely represented with common encoding structures like real arrays. Nevertheless, the presence of problem-specific constraints may require the implementation of some user-defined code to modify and evaluate candidate solutions. A relevant design aspect followed by JCLEC-MO is to minimize the need to recompile the framework if some external piece of code has to be injected. To do this, JCLEC-MO allows specifying external code dependencies in XML-based configuration files. Besides this framework, to the best of our knowledge, only ECJ and Opt4J support the edition of configuration files to reference external code. Genetic operators could be also integrated and configured by the configuration file.
    
\emph{Batch processing and parallel evaluation}.
Batch processing is a common feature of MOFs, including JCLEC-MO. It acquires greater importance in the case of industrial environments, as they usually require the execution of multiple experiments in order to validate the best choice of an algorithm for a particular complex engineering problem, and also to consider alternative inputs to carry out different studies related to the project. In the case of demanding problems requiring the execution of time-consuming algorithms, parallel processing can be a differentiating factor for the selection of a framework. Most frameworks, including JCLEC-MO, support parallel evaluation of solutions. However, more advanced features like parallel execution of experiments or parallel metaheuristics are not so extended among the analyzed frameworks yet.
    
\emph{Experimental support}.
Adapting a solution to an engineering problem is an iterative process that demands a significant effort in repeating executions, validating outcomes and adapting the framework to the specific needs. Hence, the availability of a complete toolkit with highly customizable utilities, such as quality indicators, benchmarks and reporters, may simplify this task and improve productivity. JCLEC-MO provides the most extensive collection of quality indicators and a representative set of benchmarks. In terms of continuous test functions, ECJ, jMetal and MOEA Framework offer the greatest number of alternatives, including the well-known test suites DTLZ and ZDT, also supported by JCLEC-MO. Furthermore, graphical reporters allow the visual inspection of outputs. HeuristicLab, EvA, Opt4J, jMetal and MOEA Framework enable the visualization of bi-dimensional PFs. MOEA Framework and jMetal can also depict the evolution of quality indicators along the execution. Similarly, JCLEC-MO permits computing indicators at a user-defined frequency and, because of its integration with R, relies on R the generation of any boxplot and parallel coordinate chart, the latter kind of graphic being specifically well-suited for many-objective optimization. In fact, the connection between JCLEC-MO and R opens up the flexible use of non-predefined statistical tests. In contrast, HeuristicLab and MOEA Framework include their own implementations.
    
\emph{Tool interoperability}.
We can speculate that MOFs are a complementary tool for engineers. Consequently, providing bridges to other suites would facilitate its integration and adoption in industrial environments. Besides the use of XML for configuring experiments, JCLEC-MO makes use of standard, text-based data formats like XML and CSV for inputs and outputs. For instance, the use of XML is also promoted by Opt4j, and greatly reduces the effort required to configure multiple experiments. However, the rest of MOFs apply other less structured formats, such as key-value pairs (ECJ, ParadisEO-MOEO and PISA) and YAML (EvA), or they require a specific code development using their own constructs, e.g. jMetal, MOEA Framework, HeuristicLab and PaGMO. Nevertheless, using well-defined formats enables interoperability with external tools, e.g. for data analysis purposes. JCLEC-MO is also fully compatible with datapro4j and R, a valuable factor that increases the potential of the suite. To the best of our knowledge, no other MOF provides open access to extended external functionalities.

In short, JCLEC-MO takes into consideration non-experts but still interested users in metaheuristic search looking for ready-to-use functionalities with a high customization degree. As can be observed from the aforementioned points, JCLEC-MO is also competitive with respect to MOO-specific alternatives like jMetal or MOEA Framework, while maintaining the essential characteristics of a general-purpose suite.

\section{Concluding remarks}\label{sec:conclusions}

Ongoing advances in artificial intelligence allow facing truly complex real-world optimization problems that could not be addressed without specialized software tools. In this scenario, metaheuristic optimization frameworks offer numerous advantages, as they provide diverse algorithms, multiple configuration and experimentation possibilities, such as synthetic test problems, and different sorts of utilities. Focusing on MOO, this sort of suites are intended to provide more flexibility regarding the definition of MOPs, as well as to include recent trends like many-objective optimization.

JCLEC-MO can be viewed as a Java suite that facilitates the integration and development of search-based solutions requiring the application of either multi-objective or many-objective optimization for multi-platform systems. With this aim, best practices in design and development have been taken into account to satisfy requirements related to aspects like extensibility, adaptability, transparency or interoperability, among others. Moreover, this suite provides the required support to conduct experiments and analyze their results using external libraries and languages widely used like R, a reference in terms of data analysis. We also illustrate the use of JCLEC-MO with a case study based on the WRM problem, which briefly explains how a complex many-objective problem could be solved in a simple sequence of steps.

Due to the rapid progress of the MOO field, the integration of novel techniques clearly constitutes a future line of work. Thanks to the provided extension points, we also plan to continue updating JCLEC-MO to support other metaheuristic paradigms and hybrid approaches, and to provide new utilities and integration capabilities with other languages that might be useful to engineers. Similarly, we consider interesting the possibility of making accessible the functions of JCLEC-MO as a service located in a cloud-based infrastructure in order to enable its interoperability with other programming paradigms.

\section*{Acknowledgments}
This work was supported by the Spanish Ministry of Economy and Competitiveness [project TIN2017-83445-P]; the Spanish Ministry of Education under the FPU program [grant FPU13/01466]; and FEDER funds.

\section*{Additional material}

Technical documentation of JCLEC-MO and code of the case study are available from~\url{http://www.uco.es/grupos/kdis/jclec-mo}.

\section*{References}
\bibliography{references}

\end{document}